\documentclass{article}

\usepackage{microtype}
\usepackage{graphicx}
\usepackage{subfigure}
\usepackage{caption}
\usepackage{booktabs} 

\usepackage{hyperref}
\newcommand{\eat}[1]{}
\usepackage{xspace}
\usepackage{amsmath}
\usepackage{amssymb}
\usepackage{amsthm}

\eat{
Note that publication-quality tables \emph{do not contain vertical
  rules.} We strongly suggest the use of the \verb+booktabs+ package,
which allows for typesetting high-quality, professional tables:
\begin{center}
  \url{https://www.ctan.org/pkg/booktabs}
\end{center}
This package was used to typeset Table~\ref{sample-table}.

\begin{table}[t]
  \caption{Sample table title}
  \label{sample-table}
  \centering
  \begin{tabular}{lll}
    \toprule
    \multicolumn{2}{c}{Part}                   \\
    \cmidrule{1-2}
    Name     & Description     & Size ($\mu$m) \\
    \midrule
    Dendrite & Input terminal  & $\sim$100     \\
    Axon     & Output terminal & $\sim$10      \\
    Soma     & Cell body       & up to $10^6$  \\
    \bottomrule
  \end{tabular}
\end{table}

}

\newcommand{\myvec}[1]{\mathbf{#1}}
\newcommand{\myvecsym}[1]{\boldsymbol{#1}}

\newcommand{\mysubsection}[1]{\subsection{#1}}

\newcommand{\vepsilon}{\myvecsym{\epsilon}}

\newcommand{\vgamma}{\myvecsym{\gamma}}

\newcommand{\vphi}{\myvecsym{\phi}}

\newcommand{\vpi}{\myvecsym{\pi}}

\newcommand{\vtheta}{\myvecsym{\theta}}

\newcommand{\vh}{\myvec{h}}

\newcommand{\vs}{\myvec{s}}

\newcommand{\vu}{\myvec{u}}
\newcommand{\vv}{\myvec{v}}

\newcommand{\vx}{\myvec{x}}

\newcommand{\vz}{\myvec{z}}

\newcommand{\vA}{\myvec{A}}
\newcommand{\vB}{\myvec{B}}

\newcommand{\vQ}{\myvec{Q}}
\newcommand{\vR}{\myvec{R}}

\newcommand{\mymathcal}[1]{\mathcal{#1}}

\newcommand{\calS}{\mymathcal{S}}

\def\gN{{\mathcal{N}}}

\def\gU{{\mathcal{U}}}

\usepackage{undertilde}
\usepackage{accents}

\newcommand{\Cat}{\mathrm{Cat}}

\newcommand{\gauss}{\mathcal{N}}

\newcommand{\softmax}{\calS}

\newcommand{\argmax}{\operatornamewithlimits{argmax}}

\newcommand{\real}{\mathbb{R}}

\newcommand{\KL}{\mathbb{KL}}

\newcommand{\expectQ}[2]{\mathbb{E}_{{#2}}\left[ {#1} \right]}

\newcommand{\be}{\begin{equation}}
\newcommand{\ee}{\end{equation}}
\newcommand{\bea}{\begin{eqnarray}}
\newcommand{\eea}{\end{eqnarray}}
\newcommand{\beaa}{\begin{eqnarray*}}
\newcommand{\eeaa}{\end{eqnarray*}}
\newcommand{\ba}{\begin{align*}}
\newcommand{\ea}{\end{align*}}

\usepackage{paralist} 

\eat{

}

\usepackage{cleveref}

\usepackage[accepted]{icml2020}

\icmltitlerunning{Collapsed amortized variational inference for SNLDS}

\begin{document}

\twocolumn[
\icmltitle{Collapsed amortized variational inference for \\ 
switching nonlinear dynamical systems}



\icmlsetsymbol{equal}{*}

\begin{icmlauthorlist}
\icmlauthor{Zhe Dong}{goog}
\icmlauthor{Bryan A.~Seybold}{goog}
\icmlauthor{Kevin P.~Murphy}{goog}
\icmlauthor{Hung H.~Bui}{vinai}
\end{icmlauthorlist}

\icmlaffiliation{goog}{Google AI, Mountain View, California, USA}
\icmlaffiliation{vinai}{VinAI Research, Hanoi, Vietna}

\icmlcorrespondingauthor{Zhe Dong}{zhedong@google.com}

\icmlkeywords{dynamical system, variational inference, particle filter}

\vskip 0.3in
]

\printAffiliationsAndNotice{}




\begin{abstract}
We propose an efficient inference method for switching nonlinear dynamical systems. The key idea is to learn an inference network which can be used as a proposal distribution for the continuous latent variables, while performing exact marginalization of the discrete latent variables. This allows us to use the reparameterization trick, and apply end-to-end training with stochastic gradient descent. We show that the proposed method can successfully segment time series data, including videos and 3D human pose, into meaningful ``regimes'' by using the piece-wise nonlinear dynamics. 
\end{abstract}

\section{Introduction} 

\begin{figure}[h]
\centering
\includegraphics[width=\linewidth]{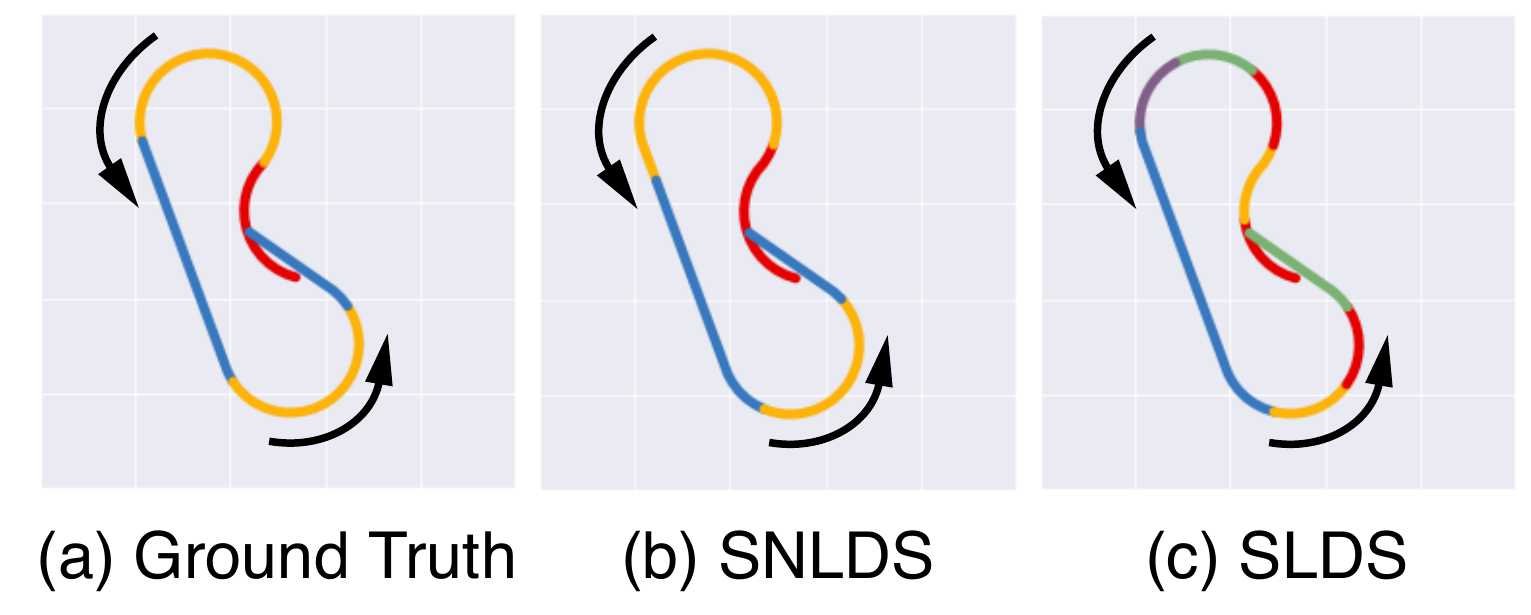}
\caption{
(a): Trajectory and ground truth segmentation of a particle. The direction of motion is indicated by the arrows. Blue is moving straight, yellow is turning counter-clockwise, red is turning clockwise.
(c) Segmentation learned by our SNLDS model.
(d) Segmentation learned by a SLDS model.
Note that to model the nonlinear dynamics, the SLDS model needs to use more segments.
}
\label{fig:dubins}
\end{figure}

Consider looking down on an airplane flying across country or a car driving through a field. The vehicle's motion is composed of straight, linear dynamics and curving, nonlinear dynamics. An example is illustrated in \cref{fig:dubins}(a). In this paper, we propose a new inference algorithm for fitting switching nonlinear dynamical systems (SNLDS), which can be used to segment time series of high-dimensional signals, such as videos, or lower dimensional signals, such as (x,y) locations, into meaningful discrete temporal ``modes'' or ``regimes''. The transitions between these modes may correspond to the changes in internal goals of the agent (e.g., a mouse switching from running to resting, as in~\citet{svae}) or may be caused by external factors (e.g., changes in the road curvature). Discovering such discrete modes is useful for scientific applications (c.f., \citet{wiltschko2015mapping, Linderman2019, sharma2018pointprocess}) as well as for planning in the context of hierarchical reinforcement learning (c.f., \citet{kipf2019compositional}).

Extensive previous work, some of which we review in \Cref{sec:related}, explores modeling temporal data using various forms of state space models (SSM). We are interested in the class of SSM which has both discrete and continuous latent variables, which we denote by $s_t$ and $\vz_t$, where $t$ is the discrete time index. The discrete state, $s_t \in \{1,2,\ldots,K\}$, represents the mode of the system at time $t$, and the continuous state, $\vz_i \in \real^H$, represents other factors of variation, such as location and velocity. The observed data is denoted by $\vx_t \in \real^D$, and can either be a low dimensional projection of $\vz_t$, such as the current location, or a high dimensional signal that is informative about $\vz_t$, such as an image. We may optionally have observed input or control signals $\vu_t \in \real^U$, which drive the system in addition to unobserved stochastic noise. We are interested in learning a generative model of the form $p_{\vtheta}(s_{1:T},\vz_{1:T},\vx_{1:T}|\vu_{1:T})$ from partial observations, namely $(\vx_{1:T}, \vu_{1:T})$. This requires inferring the posterior over the latent states, $p_{\vtheta}(s_{1:T},\vz_{1:T}|\vv_{1:T})$, where $\vv_t=(\vx_t,\vu_t)$ contains all the visible variables at time $t$. For training purposes, we usually assume that we have multiple such trajectories, possibly of different lengths, but we omit the sequence indices from our notations for simplicity. This problem is very challenging, because the model contains both discrete and continuous latent variables (a so-called ``hybrid system'') and has nonlinear transition and observation models.

\begin{figure}[t]
\centering
\includegraphics[width=0.9\linewidth]{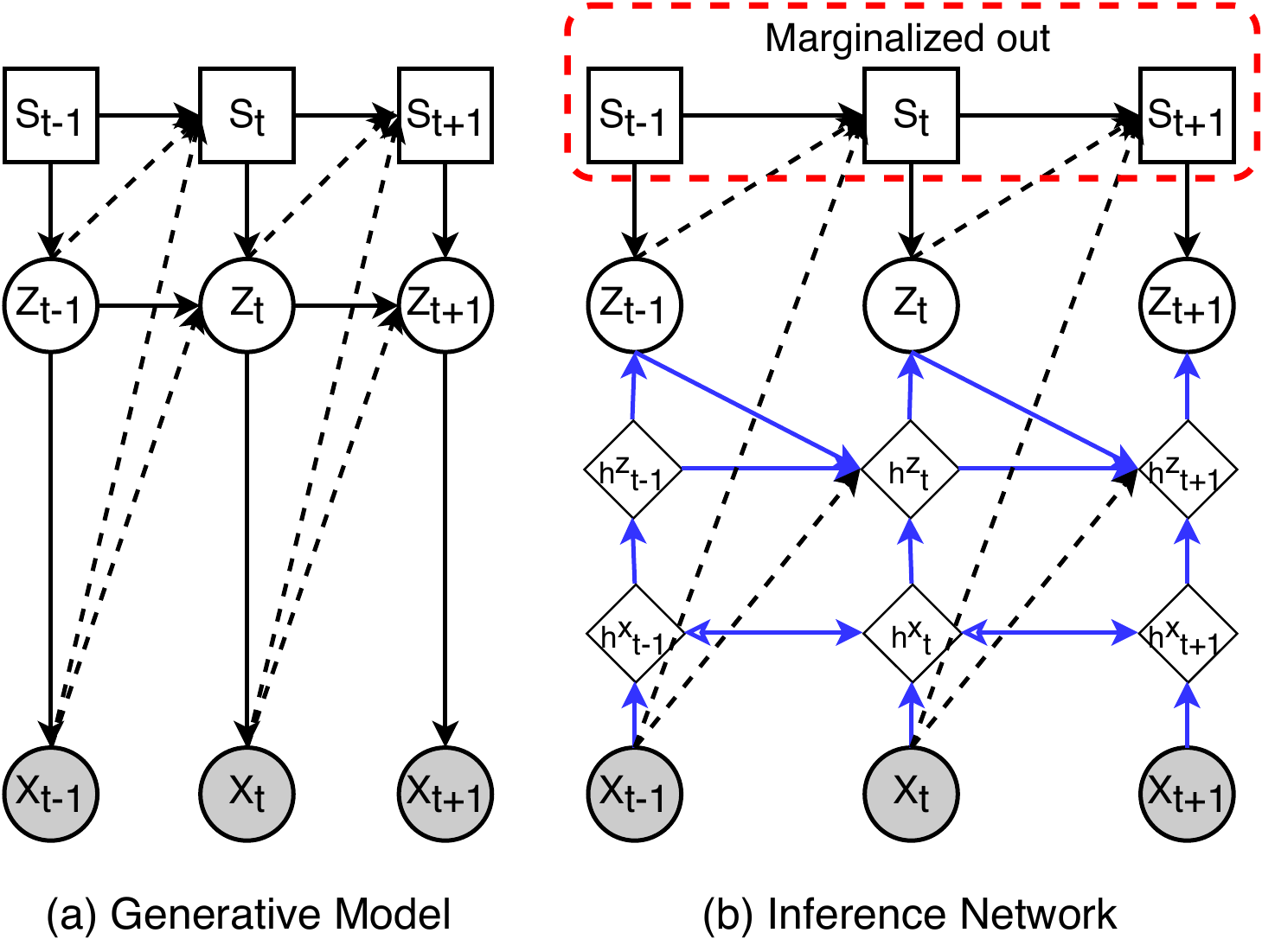}
\caption{
{\bf Left:} Illustration of the generative model. Dashed arrows indicate optional connections.
{\bf Right:} Illustration of the inference network. Solid black arrows share parameters $\vtheta$ with the generative model, solid blue arrows have parameters $\vphi$ that are unique to $q$. The diamonds represent deterministic nodes computed with RNNs: $h_t^x$ is a bidirectional RNN applied to $\vx_{1:T}$, and $h_t^z$ is a unidrectional RNN applied to $\vh_{t-1}^x$ and $\vz_{t-1}$.
}
\label{fig:model}
\end{figure}

The main contribution of our paper is a new way to perform efficient approximate inference in this class of SNLDS models. The key observation is that, conditioned on knowing $\vz_{1:T}$ as well as $\vv_{1:T}$, we can marginalize out $s_{1:T}$ in linear time using the forward-backward algorithm. In particular, we can efficiently compute the gradient of the log marginal likelihood, $\nabla \sum_{s_{1:T}} \log p(s_{1:T} | \tilde{\vz}_{1:T}, \vv_{1:T})$, where $\tilde{\vz}_{1:T}$ is a posterior sample that we need for model fitting. To efficiently compute posterior samples $\tilde{\vz}_{1:T}$, we learn an amortized inference network $q_{\vphi}(\vz_{1:T}|\vv_{1:T})$ for the ``collapsed'' NLDS model  $p(\vz_{1:T},\vv_{1:T})$. Collapsing removes the discrete variables, and allows us to use reparameterization for the continuous $\vz$. These tricks let us use stochastic gradient descent (SGD) to learn $p$ and $q$ jointly, as explained in \Cref{sec:method}. We can then use $q$ as a proposal distribution inside a Rao-Blackwellised particle filter~\citep{Doucet2000}, although in this paper, we just use a single posterior sample, as is common with Variational AutoEncoders (VAEs,~\citet{kingma2013vae, Rezende2014Vae}).

Although the above ``trick'' allows us efficiently perform inference and learning, we find that in challenging problems (e.g., when the dynamical model $p(\vz_t|\vz_{t-1},\vv_t)$ is very flexible), the model uses only a single discrete latent variable and does not perform mode switching. This is a form of ``posterior collapse'', similar to VAEs, where powerful decoders can cause the latent variables to be ignored, as explained in~\citet{brokenElbo}. Our second contribution is a new form of posterior regularization, which prevents the aforementioned problem and results in a significantly improved segmentation.

We apply our method, as well as various existing methods, to two previously proposed low-dimensional time series segmentation problems, namely a $1$d bouncing ball, and a $2$d moving arm. In the $1$d case, the dynamics are piecewise linear, and all methods perform perfectly. In the $2$d case, the dynamics are piecewise nonlinear, and we show that our method infers much better segmentation than previous approaches for comparable computational cost. We also apply our method to a simple new video dataset (see \cref{fig:dubins} for an example) and sequences of human poses, and find that it performs well, provided we use our proposed regularization method.

In summary, our main contributions are
\begin{itemize}
\setlength{\itemsep}{1pt}
\setlength{\parskip}{0pt}
\setlength{\parsep}{0pt}
\item Learning switching nonlinear dynamical systems parameterized with neural networks by marginalizing out discrete variables.
\item Using entropy regularization and annealing to encourage discrete state transitions.
\item Demonstrating that the discrete states of nonlinear models are more interpretable.
\end{itemize}

\section{Related Work}
\label{sec:related}

\mysubsection{State space models}

We consider the following state space model:
\begin{align}
p_{\theta}(\vx,\vz,\vs) &=  p(\vx_1|\vz_1) p(\vz_1|s_1) \\ \nonumber & \left[ \prod_{t=2}^T p(\vx_t|\vz_t) p(\vz_t|\vz_{t-1},s_t) p(s_t|s_{t-1}, \vx_{t-1}) \right],
\end{align}
where $s_t \in \{1,\ldots,K\}$ is the discrete hidden state, $\vz_t \in \real^L$ is the continuous hidden state, and $\vx_t \in \real^D$ is the observed output, as in \cref{fig:model}(a). For notational simplicity, we ignore any observed inputs or control signals $\vu_t$, but these can be trivially added to our model.

Note that the discrete state influences the latent dynamics $\vz_t$, but we could trivially make it influence the observations $\vx_t$ as well. More interesting are which edges we choose to add as parents of the discrete state $s_t$. We consider the case where $s_t$ depends on the previous discrete state, $s_{t-1}$, as in a hidden Markov model (HMM), but also depends on the previous observation, $\vx_{t-1}$. This means that state changes do not have to happen ``open loop'', but instead may be triggered by signals from the environment. We can trivially depend on multiple previous observations; we assume first-order Markov for simplicity. We can also condition $\vz_t$ on $\vx_{t-1}$, and $s_t$ on $\vz_{t-1}$. It is straightforward to handle such additional dependencies (shown by dashed lines in~\cref{fig:model}(a)) in our inference method, which is not true for some of the other methods we discuss below.

We still need to specify the functional forms of the conditional probability distributions. In this paper, we make the following fairly weak assumptions:
\begin{align}
  p(\vx_t|\vz_t) &= \gauss(\vx_t|f_x(\vz_t),\vR), \\
  p(\vz_t|\vz_{t-1},s_t=k) &= \gauss(\vz_t|f_z(\vz_{t-1},k),\vQ), \\
  p(s_t|s_{t-1}=j,\vx_{t-1}) &= \Cat(s_t|\softmax(f_s(\vx_{t-1}, j)),
\end{align}
where $f_{x,z,s}$ are nonlinear functions (MLPs or RNNs), $\gauss(\cdot, \cdot)$ is a multivariate Gaussian distribution, $\Cat(\cdot)$ is a categorical distribution, and $\softmax(\cdot)$ is a softmax function. $\vR \in \real^{D\times D}$ and $\vQ \in \real^{H\times H}$ are learned covariance matrices for the Gaussian emission and transition noise.

If $f_x$ and $f_z$ are both linear, and $p(s_t|s_{t-1})$ is first-order Markov without dependence on $\vz_{t-1}$, the model is called a switching linear dynamical system (SLDS). If we allow $s_t$ to depend on $\vz_{t-1}$, the  model is called a recurrent SLDS~\citep{Linderman2017,Linderman2017exploit}. We will compare to rSLDS in our experiments.

If $f_z$ is linear, but $f_x$ is nonlinear, the model is sometimes called a  ``structured variational autoencoder'' (SVAE)~\citep{svae}, although that term is ambiguous, since there are many forms of structure. We will compare to SVAEs in our experiments.

If $f_z$ is a linear function, the model may need to use many discrete states in order to approximate the nonlinear dynamics, as illustrated in~\cref{fig:dubins}(d). We therefore allow $f_z$ (and $f_x$) to be nonlinear. The resulting model is called a switching nonlinear dynamical system (SNLDS), or Nonlinear Regime-Switching State-Space Model (RSSSM)~\citep{Chow2013}. Prior work typically assumes $f_z$ is a simple nonlinear model, such as polynomial regression. If we let $f_z$ be a very flexible neural network, there is a risk that the model will not need to use the discrete states at all. We discuss a solution to this in~\Cref{sec:regularizers}.

The discrete dynamics can be modeled as a semi-Markov process, where states have explicit durations (see e.g.,~\citet{Duong2005,Chiappa2014}). One recurrent, variational version is the recurrent hidden semi-Markov model (rHSMM,~\citet{Hanjun2017rhsmm}). Rather than having a stochastic continuous variable at every timestep, rHSMM instead stochastically switches between states with deterministic dynamics. The semi-Markovian structures in this work have an explicit maximum duration, which makes them less flexible. A revised method, ~\citep{kipf2019compositional}, is able to better handle unknown durations, but produces a potentially infinite number of distinct states, each with deterministic dynamics. The deterministic dynamics of these works may limit their ability to handle noise. 

\mysubsection{Variational inference and learning}

A common approach to learning latent variable models is to maximize the evidence lower bound (ELBO) on the log marginal likelihood (see e.g.,~\citet{Blei2016review}). This is given by
$
\log p(\vx)
 \leq \mathcal{L}(\vx;\vtheta,\vphi)
  =\expectQ{\log p_{\vtheta}(\vx,\vz,\vs)-\log q_{\vphi}(\vz,\vs|\vx)}{q_{\vphi}(\vz,\vs|\vx)},
$
where $q_{\vphi}(\vz,\vs|\vx)$ is an approximate posterior.\footnote{
In the case of sequential models, we can create tighter lower bounds using methods such as FIVO~\citep{Maddison2017fivo}, although this is orthogonal to our work.} 
Rather than computing $q$ using optimization for each $\vx$, we can train an inference network, $f_{\vphi}(\vx)$, which emits the parameters of $q$. This is known as "amortized inference" (see e.g.,~\citet{kingma2013vae}).

If the posterior distribution $q_{\vphi}(\vz,\vs|\vx)$ is reparameterizable, then we can make the noise independent of $\vphi$, and hence apply the standard SGD to optimize $\vtheta,\vphi$. Unfortunately, the discrete distribution $p(\vs|\vx)$ is not reparameterizable. In such cases, we can either resort to higher variance methods for estimating the gradient, such as REINFORCE, or we can use continuous relaxations of the discrete variables, such as Gumbel Softmax~\citep{Jang2017}, Concrete~\citep{Maddison2017}, or combining both, such as REBAR~\citep{Tucker2017}. We will compare against a Gumbel-Softmax version of SNLDS in our experiments. The continuous relaxation approach was applied to SLDS models in~\citep{becker2019} and HSMM models in~\citep{liu2018structured, kipf2019compositional}. However, the relaxation can lose many of the benefits of having discrete variables \citep{RWS}. Relaxing the distribution to a soft mixture of dynamics results in the Kalman VAE (KVAE) model of~\citet{fraccaro2017kvae}. We will compare to KVAE in our experiments. A concern is that soft models may use a mixture of dynamics for distinct ground truth states rather than assigning a distinct mode of dynamics at each step as a discrete model must do. In~\Cref{sec:method}, we propose a new method to avoid these issues, in which we collapse out $\vs$ so that the entire model is differentiable. 

The SVAE model of \citet{svae} also uses the forward-backward algorithm to compute $q(\vs|\vv)$; however, they assume the dynamics of $\vz$ are linear Gaussian, so they can apply the Kalman smoother to compute $q(\vz|\vv)$. Assuming linear dynamics can result in over-segmentation, as we have discussed. A forward-backward algorithm is applied once to the discrete states and once to the continuous states to compute a structured mean field posterior $q(\vz) q(\vs)$. In contrast, we perform approximate inference for $\vz$ using one forward-backward pass of a non-linear network and then exact inference for $\vs$ using a second pass, as we explain in~\Cref{sec:method}.

\mysubsection{Monte Carlo inference}
There is a large literature on using sequential Monte Carlo methods for inference in state space models as particle filters (see e.g.,~\citet{Doucet2011pf}). When the model is nonlinear (as in our case), we may need many particles to get a good approximation, which can be expensive. We can often get better (lower variance) approximations by analytically marginalizing out some of the latent variables; the resulting method is called a ``Rao Blackwellised particle filter'' (RBPF).

Prior work (e.g.,~\citet{Doucet01jmls}) has applied RBPF to SLDS models, leveraging the fact that it is possible to marginalize out $p(\vz|\vs,\vv)$ using the Kalman filter. It is also possible to compute the optimal proposal distribution for sampling from $p(\vs_t|\vs_{t-1},\vv)$ in this case. However, this relies on the model being conditionally linear Gaussian. In contrast, we marginalize out $p(\vs|\vz,\vv)$, so we can handle nonlinear models. In this case, it is hard to compute the optimal proposal distribution for sampling from $p(\vz_t|\vz_{t-1},\vv)$, so instead we use variational inference to learn to approximate this.

\section{Method} 
\label{sec:method}

\mysubsection{Inference}

We use the following variational posterior:
$
q_{\vphi,\vtheta}(\vz,\vs|\vx) = q_{\vphi}(\vz|\vx) p_{\vtheta}(\vs|\vz,\vx),
$
where $p_{\vtheta}(\vs|\vz,\vx)$ is the exact posterior (under the generative model) computed using the forward-backward algorithm, and $q_{\vphi}(\vz|\vx)$ is defined below. To compute $q_{\vphi}(\vz|\vx)$, we first process $\vx_{1:T}$ through a bidirectional RNN, whose state at time $t$ is denoted by $\vh_t^x$. We then use a forward (causal) RNN, whose state denoted by $\vh_t^z$, to compute the parameters of $q(\vz_t|\vz_{1:t-1},\vx_{1:T})$, where the hidden state is computed based on $\vh_{t-1}^z$ and $\vh_t^x$. This gives the following approximate posterior:
$
q_{\vphi}(\vz_{1:T}|\vx_{1:T})
= \prod_t q(\vz_t|\vz_{1:t-1},\vx_{1:T})
= \prod_t q(\vz_t|\vh^z_t).
$
See~\cref{fig:model}(b) for an illustration.

We can draw a sample $\vz_{1:T} \sim q_{\vphi}(\vz|\vx)$ sequentially, and then treat this as ``soft evidence'' for the HMM model. We can use a forward-backward algorithm to integrate out the discrete variables and compute gradients as Eqn.~\ref{eqn:derivative}. This approach offers a great amount of modeling flexibility. The only constraints are that $q(\vz|\vx)$ is differentiable and that the discrete variables can be integrated out of $p(\vx,\vz)$ to also make it differentiable. The continuous transition dynamics can be linear, a simple non-linear kernel function, or a complicated function parameterized as an artificial neural network or RNN. The discrete transitions can depend on observed data, control signals, or the soft evidence samples, $\vz_{1:T}$. The flexibility of this formulation allows it to cover the model families of multiple prior works~\cite{svae,Linderman2017,Chow2013,Doucet2000} with a single core algorithm.

\mysubsection{Learning}

The evidence lower bound (ELBO) for a single sequence $\vx$ is given by
\begin{eqnarray}
\mathcal{L}_\text{ELBO}
& = & \mathbb{E}_{q_{\vphi}(\vz|\vx)p_{\vtheta}(\vs|\vx,\vz)}[ \log         p_{\vtheta}(\vx,\vz)p_{\vtheta}(\vs|\vx,\vz)  \nonumber \\
&& -\log q_{\vphi}(\vz|\vx)p_{\vtheta}(\vs|\vx,\vz) ] \\
& =& \expectQ{\log p_{\vtheta}(\vx,\vz) -\log q_{\vphi}(\vz|\vx)}{q_{\vphi}(\vz|\vx)}  
\end{eqnarray}

Because $q_{\vphi}(\vz)$ is reparameterizable, we can approximate the gradient as follows:
\begin{align}
\nabla_{\theta,\phi} \mathcal{L}(\theta,\phi)
\approx
\nabla_{\theta,\phi} 
\log p_{\theta}(\vx,\tilde{\vz})
-
\nabla_{\phi} 
\log q_{\phi}(\tilde{\vz}|\vx)
\end{align}
where $\tilde{\vz}$ is a sample from the variational proposal
$
\tilde{\vz} \sim 
q_{\vphi}(\tilde{\vz}_1|\vx_{1:T})
\prod_{t=2}^T
q_{\vphi}(\tilde{\vz}_t|\tilde{\vz}_{t-1},\vx_{1:T}).
$
The second term can be computed by applying backpropagation through time to the inference RNN. In the appendix, we show that the first term is given by
\begin{multline}
\nabla_{\theta,\phi}\log p_{\theta}(\vx,\tilde{\vz}) $=$ ~~~~~~~~~~~~~~~~~~~~~~~~~~~~~~~~~~~~ \\
\sum_{t=2}^T  \sum_{j,k} \gamma_t^2(j,k)
\nabla \left[ \log B_t(k) A_t(j,k) \right] \\
 + \sum_k \gamma_1^1(k) \nabla
\left[\log B_1(k) \pi(k) \right]
\label{eqn:derivative}
\end{multline}
where
\begin{eqnarray*}
    A_t(j,k) &=& p(s_t=j|s_{t-1}=k,\vx_{t-1}) \\
    B_t(k) &=& p(\vx_t|\vz_t) p(\vz_t|\vz_{t-1},s_t=k)~(t>1) \\
    B_1(k) &=& p(\vx_1|\vz_1) p(\vz_1|s_1=k) \\
    \gamma_t^2(j,k) &=& p(s_t=k,s_{t-1}=j | \vx_{1:T}, \vz_{1:T}) \\
    \gamma_t^1(k) &=& p(s_t=k| \vx_{1:T}, \vz_{1:T})
\end{eqnarray*}

\mysubsection{Entropy regularization and temperature annealing}
\label{sec:regularizers}

When using expressive nonlinear functions (e.g. an RNN or MLP) to model $p(\vz_t|\vz_{t-1},s_t)$, we found that the model only used a single discrete state, analogous to posterior collpase in VAEs (see e.g.,~\citet{brokenElbo}). The forward-backward algorithm causes this behavior because low-probability states are never improved. Prior work, such as~\citep{Linderman2017}, solves this problem by multi-step pretraining to ensure the model is well initialized. To encourage the model to utilize multiple states, we add an additional regularizing term to the ELBO that penalizes the KL divergence between the state posterior at each time step and a uniform prior $p_{\text{prior}}(s_t=k)=1/K$~\citep{Burke2019entropycoef}. We call this a cross-entropy regularizer:
\be
\mathcal{L}_\text{CE} = \sum_{t=1}^T
\KL(p_{\text{prior}}(s_t)|| p(s_t | \vz_{1:T}, \vx_{1:T})).
\ee
Our overall objective now becomes
\be
\mathcal{L}(\vtheta,\vphi)  = \mathcal{L}_\text{ELBO}(\vtheta,\vphi) 
- \beta \mathcal{L}_\text{CE}(\vtheta,\vphi),
\ee
where $\beta>0$ is a scaling factor. To further smooth the optimization problem, we apply temperature annealing to the discrete state transitions, as follows:
$
p(s_t=k|s_{t-1}=j,\vx_{t-1}) = \softmax(\frac{p(s_t=k|s_{t-1}=j,\vx_{t-1})}{\tau}),
$
where $\tau$ is the temperature.

At the beginning stage of training, $\beta,~\tau$ are set to large values. Doing so ensures that all states are visited, and can explain the data well. Over time, we reduce the regularizers to $0$ and temperature to $1$, according to a fixed annealing schedule. Initially, the regularization induces correlated dynamics because each state needs to be used, but annealing allows the dynamics to decorrelate (See ~\Cref{app:correlation} and c.f.,~\citet{Rose98}). The result is similar to multi-step pretraining but our approach works in a continuous end-to-end fashion.

\section{Experiments}
\label{sec:results}

In this section, we compare our method to various other methods that have been recently proposed for time series segmentation using latent variable models. Since it is hard to evaluate segmentation without labels, we use three synthetic datasets, where we know the ground truth, for quantitative evaluation but we also qualitatively evaluate the segmentation on a real world dataset.

In each case, we fit the model to the data, and then estimate the most likely hidden, discrete state at each time step, $\hat{s}_t = \argmax q(s_t|\vx_{1:T})$. Since the model is unidentifiable, the state labels have no meaning, so we post-process them by selecting the permutation over labels that maximizes the $F_1$ score across frames. The $F_1$ score is the harmonic mean of precision and recall, $2 \times \tt{precision} \times \tt{recall} / (\tt{precision} + \tt{recall})$,  where $\tt{precision}$ is the percentage of the predictions that match the ground truth states, and $\tt{recall}$ is the percentage of the ground truth states that match the predictions. We also compute the switching-point $F_1$ by only considering the frames where the ground truth state changes. This measure compliments the frame-wise $F_1$, because it measures temporal specificity.

\mysubsection{1d bouncing ball}
\label{sec:bball}

\begin{table*}[ht]
{
    \caption{Quantitative comparisons (in \% $\pm \sigma$) for segmentation on bouncing ball and reacher task. We report the $F_1$ scores in percentage with mean and standard deviation over 5 runs. (\textit{S.P.} for switching point, \textit{F.W.} for frame-wise, the best mean is in bold.) The $F_1$ score for CompILE is adapted from~\citet{kipf2019compositional}, where only switching point $F_1$ score is provided. The $F_1$ score for KVAE is computed based on taking `argmax' on the `dynamics parameter network' as described in~\citet{fraccaro2017kvae}. }
    \begin{center}
        \begin{tabular}{c|cc|cc}
        \toprule
        {\bf DATASET} 
        & \multicolumn{2}{c}{Bouncing Ball}
        & \multicolumn{2}{c}{Reacher Task} \\
        \midrule
        {\bf METRIC}
        & $F_1$ (S.P.) 
        & $F_1$ (F.W.)
        & $F_1$ (S.P.) 
        & $F_1$ (F.W.)\\
        \midrule
        {\bf SLDS (Ours)}
        & 100.
        & 100.
        & \bf{59.6 $\pm$ 3.2}
        & \bf{81.0 $\pm$ 3.4} \\
        {\bf rSLDS}
        & 100.
        & 100.
        & $47.2\pm 3.2$
        & $69.8 \pm 3.5$ \\
        {\bf SVAE}
        & 100.
        & 100.
        & $35.3\pm 2.6$
        & $62.3 \pm 4.9$ \\        
        {\bf KVAE}
        & 100.
        & 100.
        & $21.5 \pm 8.0$
        & $33.7 \pm 7.5$ \\   
        \midrule
        {\bf SNLDS (Ours)}
        & 100.
        & 100.
        & \textbf{78.1 $\pm$ 4.2}
        & \textbf{89.0 $\pm$ 2.0} \\
        {\bf Gumbel-Softmax SNLDS}
        & $97.6 \pm 1.8$
        & $93.8 \pm 4.0$
        & $5.0 \pm 8.7$
        & $14.2 \pm 9.3$ \\
        {\bf CompILE}
        & -
        & -
        & $74.3 \pm 3.3$
        & - \\   
        \bottomrule
        \end{tabular}
    \end{center}
    \label{table:bouncingball_reacher}
}
\end{table*}

\begin{figure}[t]
    \centering
    \includegraphics[width=0.9\linewidth]{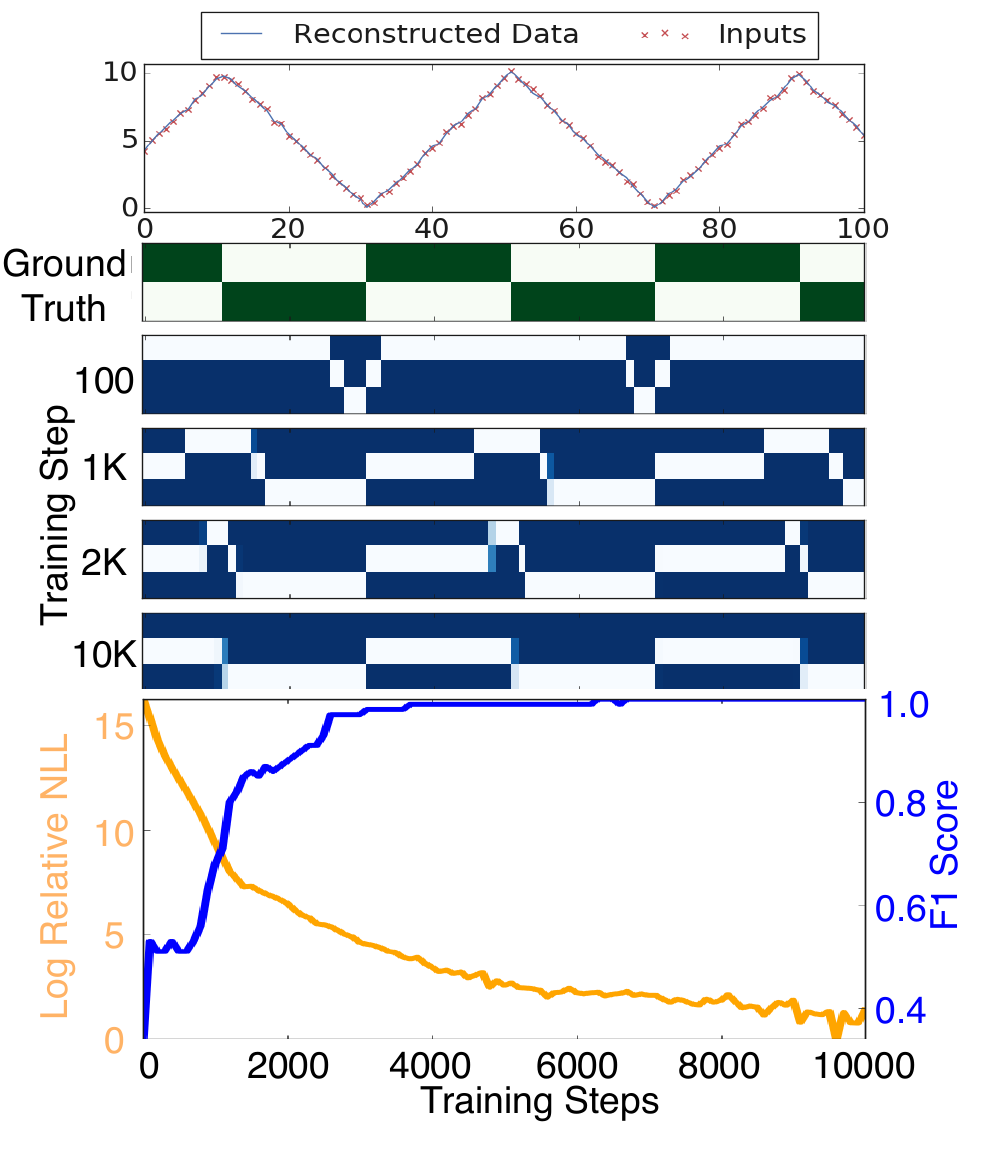}
    \caption{SNLDS Segmentation on bouncing ball task with an RNN continuous transition function. {\bf Top}: illustration of input sequence and reconstruction. {\bf Center (green)}: ground truth of the latent discrete states that correspond to the two directions of motion. {\bf Center (blue)}: the posterior marginals of $p(s_t=k| \vx_{1:T}, \vz_{1:T})$ of SNLDS at $100$, $1000$, $2000$ and $10000$ training steps, where lighter colors represent higher likelihood. {\bf Bottom}: Training progress of the log relative negative log-likelihood (Orange) and frame-wise F1  score (Blue) for SNLDS. Log relative negative log-likelihood is calculated as $\ln(\mathrm{nll} - \mathrm{min}(\mathrm{nll}) + 1.)$, where $\mathrm{nll}$ is negative log-likelihood. The scale emphasizes that the loss still improves even late during training.}
    \label{fig:bouncingball}
\end{figure}

In this section, we use a simple dataset from \citet{svae}. The data encodes the location of a ball bouncing between two walls in a one dimensional space. The initial position and velocity are random, but the wall locations are constant.

We apply our SNLDS model to this data, where $f_x$ and $f_z$ are both MLPs. We found that regularization was not necessary in this experiment. We also consider the case where $f_x$ and $f_z$ are linear (i.e. an SLDS model), the rSLDS model of~\citet{Linderman2017}, the SVAE model of~\citet{svae}, the Kalman VAE (KVAE) model of~\citet{fraccaro2017kvae} and a Gumbel-Softmax version of SNLDS as described in~\Cref{app:gs_snlds}. We use the implementations of rSLDS, SVAE, and KVAE  provided by the authors.

\begin{figure}[!htb]
  \centering
  \includegraphics[width=\linewidth]{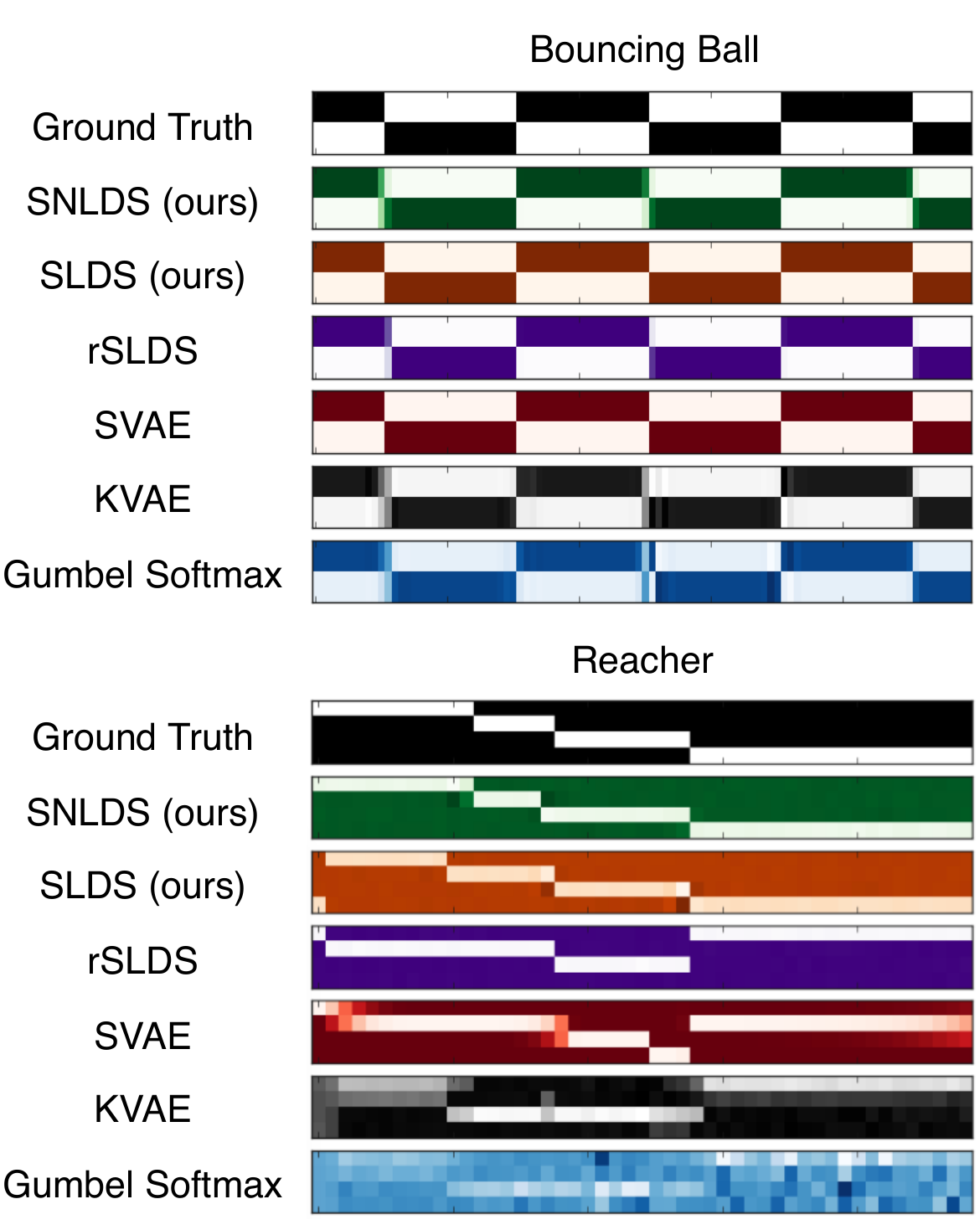}
    \caption{Segmentation on bouncing ball (top) and reacher task (bottom). From top to bottom: 
    ground truth of latent discrete states, then the posterior marginals, $p(s_t=k| \vx_{1:T}, \vz_{1:T})$, of the SNLDS, SLDS, rSLDS, SVAE, KVAE, and Gumbel-Softmax SNLDS models respectively, where lighter color represents higher probability. CompILE is not included because it represents a different model family that directly predicts the segment boundary without calculating posterior marginals at each time step.}
    \label{fig:reacher}
\end{figure}

All models we tested learn a perfect segmentation, as shown in~\Cref{fig:reacher}(a) and~\Cref{table:bouncingball_reacher}. This serves as a ``sanity check'' that we are able to use and implement the rSLDS, SVAE, KVAE and Gumbel-Softmax SNLDS code correctly. (See also~\Cref{app:bball} for further analysis.)

Note that the ``true'' number of discrete states is just 2, encoding whether the ball is moving up or down. We find that our method can learn to ignore irrelevant discrete states if they are not needed. This is presumably because we are maximizing the marginal likelihood since we sum over all hidden states, and this is known to encourage model simplicity due to the "Bayesian Occam's razor" effect~\citep{Murray05}. By contrast, we had to be more careful in setting $K$ when using the other methods.

An example of training a SNLDS model on the Bouncing Ball task is provided as Figure~\ref{fig:bouncingball}. Early in training, the discrete states do not align well to the ground truth transitions. The three states transition rapidly near one of the walls and the frame-wise F1 score is near chance values. However, by ten thousand iterations, the model has learned to ignore one state and switches between the two states corresponding to the ball bouncing from the wall. Notably the negative log-likelihood changes by over 10 orders of magnitude before the model learns accurate segmentation of even this simple problem. We hypothesize that the likelihood is dominated by errors in continuous dynamics rather than in the discrete segmentation until very late in training.

\mysubsection{2d reacher task}
\label{sec:reacher}
In this section, we consider a dataset proposed in the CompILE paper~\citep{kipf2019compositional}. The observations are sequences of $36$ dimensional vectors, derived from the 2d locations of various static objects, and the 2d joint locations of a moving arm (see~\Cref{app:reacher} for details and a visualization). The ground truth discrete state for this task is the identity of the target that the arm is currently reaching for (i.e., its "goal").

We fit the same 6 models as above to this dataset. It is a much harder problem that requires more expressive dynamics, and we found that we needed to add regularization to our model to encourage it to switch states. \Cref{fig:reacher}(b) visualizes the resulting segmentation (after label permutation) for a single example. We see that our SNLDS model matches the ground truth more closely than our SLDS model, as well as the rSLDS, SVAE, KVAE, and Gumbel-Softmax baselines.

To compare performance quantitatively, we evaluate the models from $5$ different training runs on the same held-out dataset of size $32$, and compute the $F_1$ scores. We also report the $F_1$ number from CompILE. The CompILE paper uses an iterative segmentation scheme that can detect state changes, but it does infer what the current latent state is, so we cannot include it in~\Cref{fig:reacher}(b). In~\Cref{table:bouncingball_reacher}, we find that our SNLDS method is significantly better than the other approaches.

\mysubsection{Dubins path}
\label{sec:dubins}
\begin{table*}[h]
{
    \caption{Quantitative comparisons (in \%) for S(N)LDS on Dubins path.
    For SLDS, $F_1$ scores with both greedy $1$-to-$1$ matching (\textit{Greedy}) and optimal merging (\textit{Merging}) are provided. The switching point $F_1$ scores are estimated with both precise matching (\textit{Tol 0}) or allowing at most $5$-step displacement (\textit{Tol 5}).}
    \begin{center}
        \begin{tabular}{c|ccc}
            \toprule
            {\bf METRIC}
            &{\bf SLDS (Greedy)}  
            &{\bf SLDS (Merging)}  
            &{\bf SNLDS}
            \\ 
            \midrule
            $F_1$ (Switching point, Tol 0) & $3.5 \pm 1.0$  & $4.4 \pm 3.1$ & \textbf{11.3 $\pm$ 5.7}
            \\
            $F_1$ (Switching point, Tol 5) & $33.7 \pm 2.5$ & $67.0 \pm 3.4$ & \textbf{82.5 $\pm$ 1.9}
            \\
            $F_1$ (Frame-wise) & $29.4 \pm 3.6$ & $61.5 \pm 8.0$  & \textbf{84.3 $\pm$ 7.2}
            \\ \bottomrule
        \end{tabular}
    \end{center}
    \label{table:dubins}
}
\end{table*}
In this section, we apply our method to a new dataset that is created by rendering a point moving in the 2d plane. The motion follows the Dubins model\footnote{
    \url{https://en.wikipedia.org/wiki/Dubins_path}
}, %
a simple model for piece-wise nonlinear (but smooth) motion that is commonly used in the fields of robotics and control theory because it corresponds to the shortest path between two points that can be traversed by wheeled robots, airplanes, etc. In the Dubins model, the change in direction is determined by an external control signal $u_t$. We replace this with three latent discrete control states: go straight, turn left, and turn right. These correspond to fixed, but unobserved, input signals $u_t$ (see~\Cref{appendix:dubins_path} for details). After generating the motion, we create a series of images, where we render the location of the moving object as a small circle on a white background. Our goal in generating this dataset was to assess how well we can recover latent dynamics from image data in a very simple, yet somewhat realistic, setting.

The publicly released code for rSLDS and SVAE does not support high dimensional inputs like images (even though the SVAE has been applied to an image dataset in \citet{svae}), and there is no public code for CompILE. Therefore we could not compare to these methods for this experiment. As we already showed in~\Cref{sec:reacher} that our method is much better than these other approaches, as well as Kalman VAE and Gumbel-Softmax version of SNLDS, on other tasks, we expect the same conclusion to hold on the harder task of segmenting videos.

Instead we focus on comparing inferred SNLDS states with SLDS states to determine the advantage of allowing each regime to be represented by a nonlinear model. The results of segmenting one sequence with these models using $5$ states are shown in~\Cref{fig:dubins}. We see that the SLDS model has to approximate the left and right turns with multiple discrete states, whereas the non-linear model learns a more interpretable representation.

We again compare the models in~\Cref{table:dubins} using $F_1$ scores. Since matching the exact time of the switching point is very hard in the unsupervised setting with noisy observations, we also report an F1 computed with a tolerance of detecting a change within 5 frames. Because the SLDS model used too many states, we calculated two versions of the metrics. The first was a greedy approach that optimally assigned the best single state to match each ground truth state. The second used an oracle to optimally merge states to match the ground truth. The SNLDS model significantly outperforms the SLDS in both scenarios.

\begin{figure*}[!htb]
  \centering
  \includegraphics[width=0.8\linewidth]{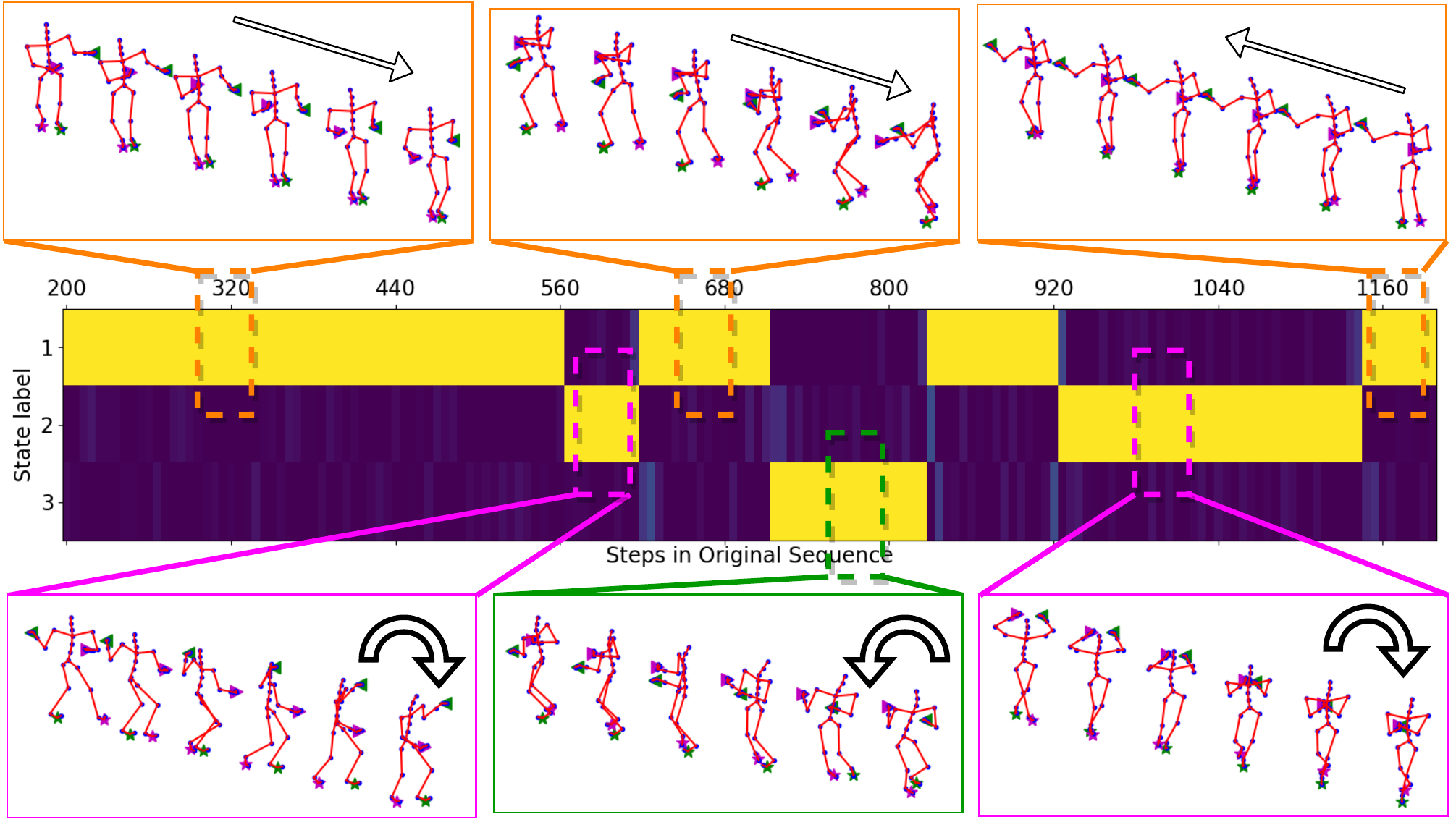}
    \caption{SNLDS segmentation result for Salsa dancing trial in CMU MoCap dataset. The model segment the motion into three different dynamical modes: moving forward and backward (orange colored), clockwise turning (magenta) and counter-clockwise turning (green). The center depicts the posterior marginal for each state and the boxes show samples of motion from each state.}
    \label{fig:salsa}
\end{figure*}

\mysubsection{Salsa Dancing from CMU MoCap}
In this section, we demonstrate the capacity of SNLDS on segmenting 3D human pose dynamics on CMU MoCap data \footnote{http://mocap.cs.cmu.edu/}. There are $30$ trials of Salsa dancing sequences in the dataset. We use $29$ of them as the training data, and hold out the other for evaluation. The training sequences are generated by down-sampling the original sequences using every 6 frames. The input to the model consists of $3D$ cordinates of $31$ joints. Using MLP to describe the nonlinear transition of continuous hidden states, SNLDS can segment sequences into $3$ modes of primitive motions, which could be interpreted as: turning clockwise, turning counter-clockwise, and translational motion. Without ground truth segmentation, we only evaluate the segmentation qualitatively, as shown in the \Cref{fig:salsa}.

\mysubsection{Analysis of the annealing schedule}
\label{sec:annealing}
Many latent variable models are trained in multiple stages to avoid getting stuck in bad local optima. For example, to fit the rSLDS model, \citet{Linderman2017} first pretrain an AR-HMM and SLDS model, and then merge them; similarly, to fit the SVAE model, \citet{svae} first train with a single latent state and then increase $K$.

We found a similar strategy was necessary for the Reacher, Dubins, and Salsa tasks, but we do this in a smooth way using annealed regularization. Early in training, we train with large temperature $\tau$ and entropy coefficient $\beta$. This encourages the model to use all states equally, so that the dynamics, inference, and emission sub-networks stabilized before beginning to learn specialized behavior. We then anneal the entropy coefficient to $0$, and the temperature to $1$ over time. We found it best to first decay the entropy coefficient $\beta$ and then decay the temperature $\tau$.

\begin{figure}[!htb]
\centering
\includegraphics[width=\linewidth]{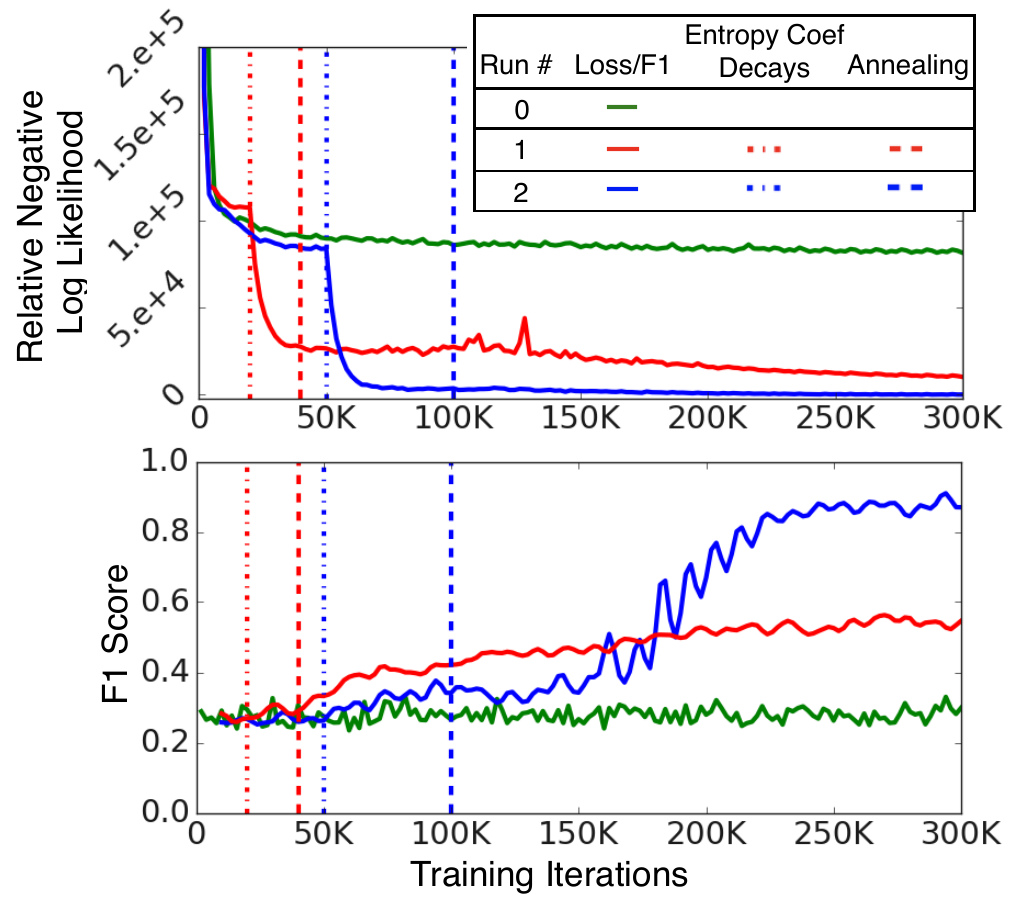}
\caption{Comparing the relative negative log-likelihood ({\bf top}) and the frame-wise $F_1$ scores ({\bf bottom}) on Dubins paths with 3 different annealing schedules. In the first run (green), the regularization coefficient and temperature start to decay at the very beginning of training. In the second run (red), the cross entropy regularization coefficient starts to decay at step $20,000$, while temperature annealing starts at step $40,000$. In the third run (blue), the coefficient decay starts at step $50,000$, while temperature annealing starts at step $100,000$. }
\label{fig:annealing}
\end{figure}

Figure~\ref{fig:annealing} demonstrates the effect of $3$ different annealing schedules on the relative log likelihood (defined as $L_t - L_{\text{min}}$, where $L_{\text{min}} = \min_t L_{t;1,2,3}$ across all three runs, and $L_t$ is the negative log-likelihood.), and the $F_1$ score. 
We find that the final negative log-likelihood and $F_1$ scores improve when we delay the annealing schedule to $50k$ steps on the Dubins task. Surprisngly, the $F_1$ score does not improve significantly until an additional $50k$ steps after the temperature begins annealing. On real problems, where we have no ground truth, we cannot use the $F_1$ score as a metric to determine the best annealing schedule. However, it seems that the schedules that improve $F_1$ the most also improve likelihood the most.

\section{Conclusion}

We have demonstrated that our proposed method can effectively learn to segment high dimensional sequences into meaningful discrete regimes. Future work includes applying this to harder image sequences and to hierarchical reinforcement learning.

\bibliography{reference}
\bibliographystyle{icml2020}

\appendix
\clearpage
\section{Appendix}
\label{sec:appendix}

\subsection{Derivation of the gradient of the ELBO}
\label{app:gradient}

The evidence lower bound objective (ELBO) of the model is defined as:
\begin{eqnarray}
\mathcal{L}
& = & \expectQ{\log p_{\theta}(\vx,\vz,\vs)-\log q_{\theta,\phi}(\vz,\vs|\vx)}{q_{\theta,\phi}(\vz,\vs|\vx)} \nonumber\\
& = & \mathbb{E}_{q_{\phi}(\vz|\vx)p_{\theta}(\vs|\vx,\vz)}[\log p_{\theta}(\vx,\vz)p_{\theta}(\vs|\vx,\vz) \nonumber \\
&&\quad \quad  \quad   \qquad  \qquad -\log q_{\phi}(\vz|\vx)p_{\theta}(\vs|\vx,\vz)] \nonumber\\
& = & \expectQ{\log p_{\theta}(\vx,\vz)}{q_{\phi}(\vz|\vx)} +H(q_{\phi}(\vz|\vx)) 
\label{equation:elbo_appendix}
\end{eqnarray}
where the first term is the model likelihood, 
and the second is the conditional entropy for variational posterior of continuous hidden states.
We can approximate the entropy of $q_{\phi}(\vz|\vx)$ as:
\begin{align*}
H(q_{\phi}(\vz|\vx)) = H(q_{\phi}(\vz_1)) + \sum_{t=2}^T H(q_{\phi}(\vz_t|\tilde{\vz}_{1:t-1}))
\end{align*}
where $\tilde{\vz}_t \sim q(\vz_t)$ is a sample from the variational posterior.
In other words, 
we compute the marginal entropy for the output of the RNN inference network at each time step,
and then sample a single latent vector to update the RNN state for the next step.

In order to apply stochastic gradient descent for end-to-end training, 
the minibatch gradient for the first term in the ELBO (Eq.~\ref{equation:elbo_appendix}) with respect to $\vtheta$
is estimated as
\begin{align*}
  \nabla_{\theta}  \expectQ{\log p_{\theta}(\vx,\vz)}{q_{\phi}(\vz|\vx)}
  =  \expectQ{\nabla_{\theta}   \log p_{\theta}(\vx,\vz)}{q_{\phi}(\vz|\vx)}
\end{align*}
For the gradient with respect to $\vphi$, 
we can use the reparameterization trick to write
\begin{align*}
 & \nabla_{\phi}  \expectQ{\log p_{\theta}(\vx,\vz)}{q_{\phi}(\vz|\vx)} \nonumber \\
 &  =  \expectQ{\nabla_{\phi}   \log p_{\theta}(\vx,\vz_{\phi}(\vepsilon,\vx))}{\vepsilon \sim N}  
\end{align*}
Therefore, the gradient is expressed as:
\begin{align*}
\nabla_{\theta} \mathcal{L} & =  
\expectQ{\nabla_{\theta} \log p_{\theta}(\vx,\vz)}{q_{\phi}(\vz|\vx)}, \\
\nabla_{\phi} \mathcal{L} & =
\expectQ{\nabla_{\phi} \log p_{\theta}(\vx,\vz_{\phi}(\vepsilon,\vx))}{\vepsilon \sim N}
+ \nabla_{\phi} H(q_{\phi}(\vz|\vx)).
\end{align*}

To compute the derivative of the log-joint likelihood
$\nabla_{\vtheta,\vphi} \log p_{\theta}(\vv)$,
where we define $\vv=(\vx_{1:T},\vz_{1:T})$ as the visible variables for brevity.
Therefore
\begin{align*}
  \nabla \log p(\vv)
=&
  \expectQ{\nabla \log p(\vv)}{p(\vs|\vv)} \\
  =&
  \expectQ{\nabla \log p(\vv,\vs)}{p(\vs|\vv)} \\
  &-\expectQ{\nabla \log p(\vs|\vv)}{p(\vs|\vv)}
  \\
  =&   \expectQ{\nabla \log p(\vv,\vs)}{p(\vs|\vv)} - 0
\end{align*}
where we used the fact that
$\log p(\vv) = \log p(\vv,\vs) - \log p(\vs|\vv)$ 
and
\begin{align*}
\expectQ{\nabla \log p(\vs|\vv)}{p(\vs|\vv)}
& = \int p(\vs|\vv)  \frac{\nabla p(\vs|\vv)}{p(\vs|\vv)} \\
& = \nabla \int p(\vs|\vv) = \nabla 1 = 0.
\end{align*}

For $\nabla \log p(\vv,\vs)$, 
we use the Markov property to rewrite it as:
\begin{align*}
&\nabla  \log p(\vv,\vs) = \\
&   \sum_{t=2}^T \nabla \log p(\vx_t|\vz_t) p(\vz_t|\vz_{t-1},s_t) p(s_t|s_{t-1}, \vx_{t-1})\\
& \qquad +\nabla \log p(\vx_1|\vz_1) p(\vz_1|s_1) p(s_1),
\end{align*}
with the expectation being:
\begin{align*}
& \nabla \log p(\vv) =  \expectQ{\nabla  \log p(\vv,\vs)}{p(\vs|\vv)} 
\\
& = \sum_k p(s_1=k|\vv) \nabla \log p(\vx_1|z_1) p(\vz_1|s_1=k) p(s_1=k)
\\
& + \sum_{t=2}^T  \sum_{j,k} p(s_{t-1}=j,s_t=k|\vv) \nabla 
\left[\log p(\vx_t|\vz_t) \cdot \right.\\
&\qquad \qquad \left.p(\vz_t|\vz_{t-1},s_t=k) p(s_t=k|s_{t-1}=j, \vx_{t-1})\right]
\\
&= \sum_{t=2}^T  \sum_{j,k} \gamma_t^2(j,k)
\nabla \log B_t(k) A_t(j,k) \\
& \qquad \qquad + \sum_k \gamma_1^1(k) \nabla \log B_1(k) \pi(k).
\end{align*}
Therefore we reach the Eq.~\ref{eqn:derivative}.

In summary, one step of stochastic gradient ascent for the ELBO can be implemented as~\Cref{alg:snlds}.

\begin{algorithm}[!h]
  \caption{SVI for Training SNLDS}
  \begin{algorithmic}[1]
    \STATE Compute $\vh_t^x$ from $\vx_{1:T}$ using a Bi-RNN;
    \STATE Recursively sample $\vz_t \sim q(\vz_t|\vz_{t-1},\vx_{1:T})$ using RNN over $\vz_{t-1}$ and $\vh_t^x$;
    \STATE Run forward-backward messages to compute $\vA$, $\vB$, $\vpi$, $\vgamma_{1:T}^1$, $\vgamma_{1:T-1}^2$ from $(\vx, \vz)$;
    \STATE Compute  $\nabla_{\vtheta,\vphi}  \log p(\vx,\vz)$ from Eqn.~\ref{eqn:derivative};
    \STATE Take gradient step.
  \end{algorithmic}
  \label{alg:snlds}
\end{algorithm}

\subsection{Gumbel-Softmax SNLDS}
\label{app:gs_snlds}
Instead of marginalizing out the discrete states with the forward-backward algorithm, one could use a continuous relaxation via reparameterization, e.g. the Gumbel-Softmax trick~\citep{Jang2017}, to infer the most likely discrete states. We call this \texttt{Gumbel-Softmax SNLDS}. 

We consider the same state space model as SNLDS:
\begin{align*}
p_{\theta}(\vx,\vz,\vs) &=  p(\vx_1|\vz_1) p(\vz_1|s_1) 
\\
&  \left[ 
  \prod_{t=2}^T p(\vx_t|\vz_t) p(\vz_t|\vz_{t-1},s_t) p(s_t|s_{t-1}, \vx_{t-1})
  \right],
\end{align*}
where $s_t \in \{1,\ldots,K\}$ is the discrete hidden state, $\vz_t \in \real^L$ is the continuous hidden state, and $\vx_t \in \real^D$ is the observed output, as in \Cref{fig:model}(a). The inference network for the variational posterior now predicts both $\vs$ and $\vz$ and is defined as
\be
q_{\vphi_\vz,\vphi_\vs}(\vz,\vs|\vx)
= q_{\vphi_\vz}(\vz|\vx) q_{\vphi_\vs}(\vs|\vx)
\ee
where
\begin{align*}
q_{\vphi_\vz}(\vz_{1:T}& |\vx_{1:T}) 
= \prod_t q(\vz_t|\vz_{1:t-1},\vx_{1:T}) \\
& = \prod_t q(\vz_t|\vh_t) \delta(\vh_t | f_{RNN}(\vh_{t-1}, \vz_{t-1}, \vh_t^{b})) \\
q_{\vphi_\vs}(\vs_{1:T}& |\vx_{1:T}) 
= \prod_t q(s_t|s_{t-1},\vx_{1:T}) \\
& = \prod_t q_{\mathrm{Gumbel-Softmax}}(s_t|g(\vh^b_t, s_{t-1}), \tau) 
\end{align*}
where $\vh_t$ is the hidden state of a deterministic recurrent neural network, $f_{RNN}(\cdot)$,
which works from left ($t=0$) to right ($t=T$),
summarizing past stochastic $\vz_{1:t-1}$.
We also feed in $\vh_t^b$, which is a bidirectional RNN, 
which summarizes $\vx_{1:T}$.
The Gumbel-Softmax distribution $q_{\mathrm{Gumbel-Softmax}}$ takes the output of a feed-forward network $g(\cdot)$ and a softmax temperature $\tau$, which is annealed according to a fixed schedule. 

The evidence lower bound (ELBO) could be written as
\begin{align}
\mathcal{L}_\text{ELBO}(\vtheta,\vphi)
& =  \mathbb{E}_{q_{\vphi_\vz}(\vz|\vx)q_{\vphi_\vs}(\vs|\vx)} 
\left[\log p_{\vtheta}(\vx,\vz,\vs)\right. \nonumber \\ 
& \qquad \left.-\log q_{\vphi_\vz}(\vz|\vx)q_{\vphi_\vs}(\vs|\vx)\right] 
\label{eqn:gs_snlds_elbo}
\end{align}
One step of stochastic gradient ascent for the ELBO can be implemented as~\Cref{alg:gumbel_softmax_snlds}.
\begin{algorithm}[!h]
  \caption{SVI for Training Gumbel-Softmax SNLDS}
  \begin{algorithmic}[1]
    \STATE Use Bi-RNN to compute $\vh_t^x$ from $\vx_{1:T}$;
    \STATE Recursively sample $\vz_t \sim q(\vz_t|\vz_{t-1},\vx_{1:T})$ using RNN over $\vz_{t-1}$ and $\vh_t^x$;
    \STATE Recursively sample $s_t$ with distribution  $q_{\mathrm{Gumbel-Softmax}}(s_t|g(\vh^b_t, s_{t-1}), \tau)$, where $g$ is a feedforward network;
    \STATE Compute the likelihood for~\cref{eqn:gs_snlds_elbo};
    \STATE Take gradient step.
  \end{algorithmic}
  \label{alg:gumbel_softmax_snlds}
\end{algorithm}

\subsection{Details on the bouncing ball experiment}
\label{app:bball}
The input data for bouncing ball experiment is a set of $100000$ sample trajectories, each of which is of $100$ timesteps with its initial position randomly placed between two walls separated by a distance of $10.$. The velocity of the ball for each sample trajectory is sampled from $\gU([-0.5, 0.5])$. The exact position of ball is obscured with Gaussian noise $\gN(0, 0.1)$. The training is performed with batch size $32$. The evaluation is carried on a fixed, held-out subset of the data with $200$ samples. For the inference network, the bi-directional and forward RNNs are both $16$ dimensional GRU. The dimensions of discrete and continuous hidden state are set to be $3$ and $4$. For SLDS, we use linear transition for continuous states. For SNLDS, we use GRU with $4$ hidden units followed by linear transformation for continuous state transition. The model is trained with fixed learning rate of $10^{-3}$, with the Adam optimizer~\citep{Kingma2015adam}, and gradient clipping by norm of $5.$ for $10000$ steps.

\subsection{Details on the reacher experiment}
\label{app:reacher}
\begin{figure}[h]
  \centering
  \includegraphics[width=\linewidth]{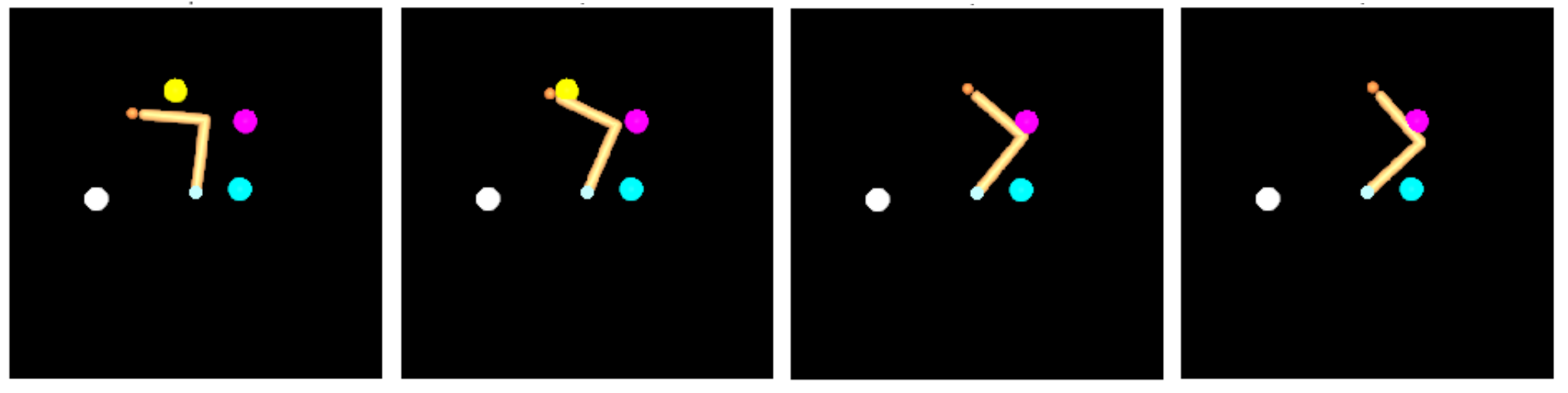}
\caption{Illustration of the observations in reacher experiment. This is $2$-D rendering of the observational vector, but the inputs to the model are sequences of vectors, as in \citet{kipf2019compositional}, not images.
}
\label{fig:reacher_ill}
\end{figure}
The observations in the reacher experiment are sequences of $36$ dimensional vectors, as described in \citet{kipf2019compositional}. First $30$ elements are the target indicator, $\alpha$, and location, $x,y$, for 10 randomly generated objects. $3$ out of $10$ objects start as targets, $\alpha=1$. The $(x,y)$ location for 5 of the non-target objects are set to $(0,0)$. A deterministic controller moves the arm to the indicated target objects. Once a target is reached, the indicator is set to $\alpha=0$. (Depicted as the yellow dot disappearing in Figure~\ref{fig:reacher_ill}.) The remaining $6$ elements of the observations are the two angles of reacher arm and the positions of two arm segment tips. The training dataset consists of $10000$ observation samples, each $50$ timesteps in length. 

This more complex task requires more careful training. The learning rate schedule is a linear warm-up, $10^{-5}$ to $10^{-3}$ over $5000$ steps, from followed by a cosine decay, with decay rate of $2000$ and minimum of $10^{-5}$. Both entropy regularization coefficient starts to exponentially decay after $50000$ steps, from initial value $1000$ with a decay rate $0.975$ and decay steps $500$. The temperature annealing follows the same exponential but only starts to decay after $100000$ steps. The training is performed in minibatches of size $32$ for $300000$ iterations using the Adam optimizer~\citep{Kingma2015adam}.

The model architecture is relatively generic. The continuous hidden state $z$ is $8$ dimensional. The number of discrete hidden states is set to $5$ for training, which is larger than the ground truth $4$ (including states targeting $3$ objects and a finished state). The observations pass through an encoding network with two $256$-unit ReLU activated fully-connected nets, before feeding into RNN inference networks to estimate the posterior distributions $q(z_t | x_{1:T})$. The RNN inference networks consist of a $32$-unit bidirection LSTM and a $64$-unit forward LSTM. The emission network is a three-layer MLP with $[256, 256, 36]$ hidden units and ReLU activation for first two layers and a linear output layer. Discrete hidden state transition network takes two inputs: the previous discrete state and the processed observations. The observations are processed by the encoding network and a $1$-D convolution with $2$ kernels of size $3$. The transition network outputs a $5 \times 5$ matrix for transition probability $p(s_t | s_{t-1})$ at each timestep. For SNLDS, we use a single-layer MLP as the continuous hidden state transition functions $p(z_t | z_{t-1}, s_t)$, with $64$ hidden units and ReLU activation. For SLDS, we use linear transitions for the continuous state.

\subsection{Details on the Dubins path experiment}
\label{app:dubins}
\label{appendix:dubins_path}

The Dubins path model\footnote{
    \url{https://en.wikipedia.org/wiki/Dubins_path}
}
is a simplified flight, or vehicle, trajectory that is the shortest path to reach a target position, given the initial position $(x_0, y_0)$, the direction of motion $\theta_0$, the speed constant $V$, and the maximum curvature constraint $\dot{\theta} \le u$. The possible motion along the path is defined by 
$$
\dot{x}_t = V \cos(\theta_t), ~
\dot{y}_t = V \sin(\theta_t), ~
\dot{\theta}_t = u.
$$
The path type can be described by three different modes/regimes: `right turn (R)' , `left turn (L)' or `straight (S).'

To generate a sample trajectory used in training or testing, we randomly sample the velocity from a uniform distribution $V \sim \gU([0.1, 0.5])$ (pixel/step), angular frequency from a uniform distribution $u/2\pi \sim \gU([0.1, 0.15])$ (/step), and initial direction $\theta_0 \sim \gU([0, 2\pi))$. The generated trajectories always start from the center of image $(0, 0)$. The duration of each regime is sampled from a Poisson distribution with mean $25$ steps, with full sequence length $100$ steps. The floating-point positional information is rendered onto a $28 \times 28$ image with Gaussian blurring with $0.3$ standard deviation to minimize aliasing. 

The same schedules as in the reacher experiment are used for the learning rate, temperature annealing and regularization coefficient decay.

The network architecture is similar to the reacher task except for the encoder and decoder networks. Each observation is encoded with a CoordConv \citep{uberai2018coordconv} network before passing into RNN inference networks, the archicture is defined in \Cref{table:coordconv_encoder}. The emission network $p(\vx_t|\vz_t)$ also uses a CoordConv network as described in \Cref{table:coordconv_decoder}. The continuous hidden state $z$ in this experiment is $4$ dimensional. The number of discrete hidden states $s$ is set to be $5$, which is larger than ground truth $3$. The inference networks are a $32$-unit bidirection LSTM and a $64$-unit forward LSTM. The discrete hidden state transition network takes the output of observation encoding network in the same manner as the reacher task. For SNLDS, we use a two-layer MLP as continuous hidden state transition function $p(z_t | z_{t-1}, s_t)$, with $[32, 32]$ hidden units and ReLU activation. For SLDS, we use linear transition for continuous states.

\begin{table*}[bth]
\centering
\caption{CoordConv encoder Architecture. Before passing into the following network, the image is padded from $[28, 28, 1]$ to $[28, 28, 3]$ with the pixel coordinates.}
\setlength\tabcolsep{5pt}
	\begin{tabular}{|c|c|c|c|c|c|}	
	\hline
	\shortstack{Layer} & Filters & Shape & Activation & Stride & Padding \\
	\hline
	1 & 2 & [5, 5] & relu & 1 & same \\
	\hline
	2 & 4 & [5, 5] & relu & 2 & same \\
	\hline
	3 & 4 & [5, 5] & relu & 1 & same \\
	\hline
	4 & 8 & [5, 5] & relu & 2 & same \\
	\hline
	5 & 8 & [7, 7] & relu & 1 & valid \\
	\hline
	6 & 8 & 2 (Kernel Size) & None & 1 & causal \\
    \hline
	\end{tabular}
\vspace{0.1cm}
\label{table:coordconv_encoder}
\vspace{-0.1cm}
\end{table*}		

\begin{table*}[bth]
\centering
\caption{CoordConv decoder Architecture. Before passing into the following network, the input $z_t$ is tiled from $[8]$ to $[28, 28, 8]$, where $8$ is the hidden dimension, and is then padded to $[28, 28, 10]$ with the pixel coordinates. }
\setlength\tabcolsep{5pt}
	\begin{tabular}{|c|c|c|c|c|c|}	
	\hline
	\shortstack{Layer} & Filters & Shape & Activation & Stride & Padding \\
	\hline
	1 & 14 & [1, 1] & relu & 1 & valid \\
	\hline
	2 & 14 & [1, 1] & relu & 1 & valid \\
	\hline
	3 & 28 & [1, 1] & relu & 1 & valid \\
	\hline
	4 & 28 & [1, 1] & relu & 1 & valid \\
	\hline
	5 & 1 & [1, 1] & relu & 1 & same \\
	\hline
	\end{tabular}
\vspace{0.1cm}
\label{table:coordconv_decoder}
\vspace{-0.1cm}
\end{table*}

\begin{figure}[h]
\centering
  \includegraphics[width=\linewidth]{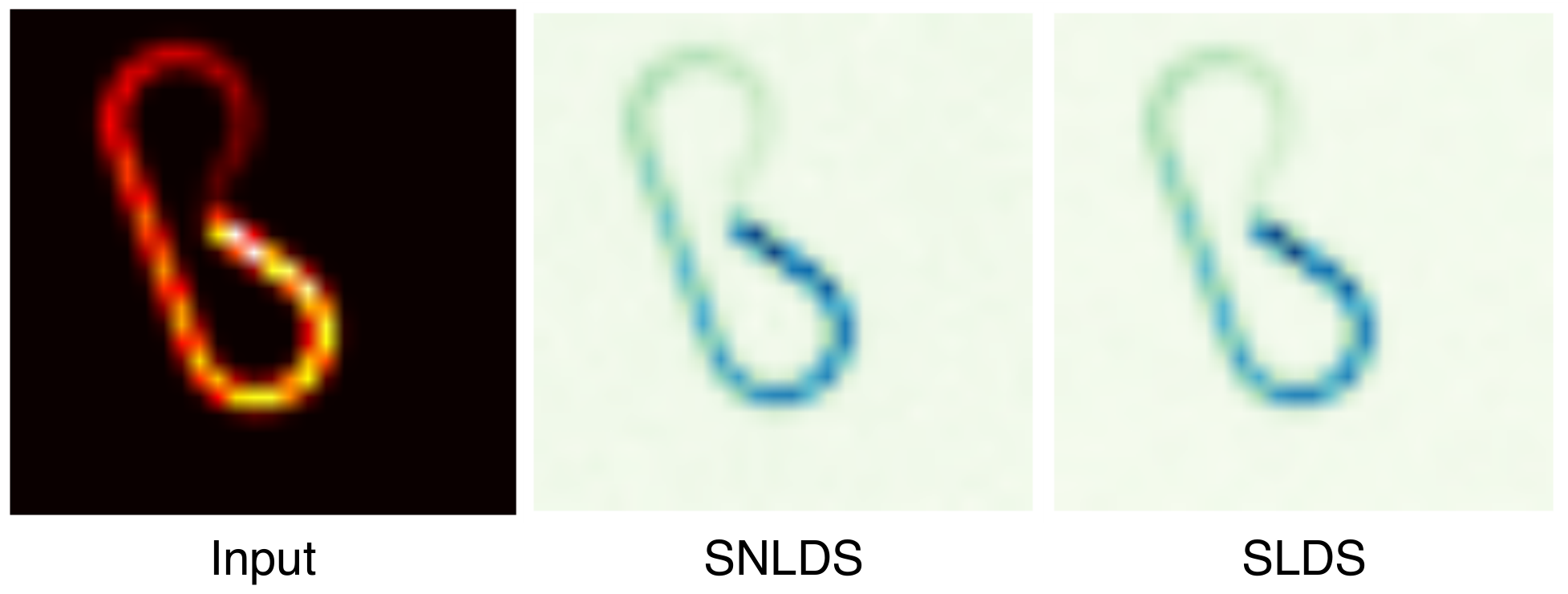}
\caption{Image sequence reconstruction for Dubins path. The sequence is averaged with early timepoints scaled to low intensity, late timepoints unchanged to indicate direction.
}
\label{fig:dubinsRecon}
\end{figure}

See \Cref{fig:dubinsRecon} for an illustration of the reconstruction abilities (of the observed images) for the SLDS and SNLDS models. They are visually very similar; however, the SNLDS has a more interpretable latent state as described in Section~\ref{sec:dubins}. 

\subsection{Regularization and Multi-steps Training}\label{app:correlation}

\begin{figure}[!htb]
\centering
  \includegraphics[width=\linewidth]{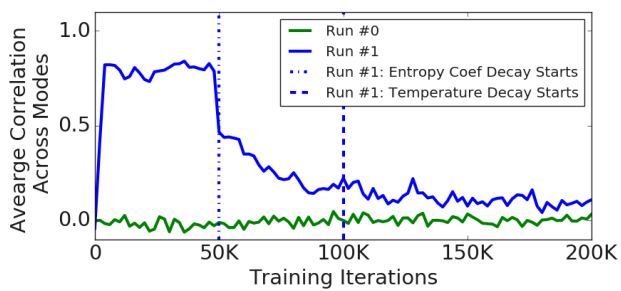}
\caption{Comparing the average Pearson correlations among the weights from individual dynamical transition modes, $p(\vz_t|\vz_{t-1},s_t=k)$, trained on Dubins Paths. Run $0$ (green) is trained without regularization. Run $1$ (blue) has its entropy coefficient starting to exponentially decay at step $50,000$, and the temperature starting to anneal at step $100,000$.}
\label{fig:correlation}
\end{figure}

Training our SNLDS model with a powerful transition network but without regularization will fit the dynamics $p(\vz_t | \vz_{t-1}, s_t)$ with a single state. With randomly initialized networks, one state fits the dynamics better at the beginning and the forward-backward algorithm will cause more gradients to flow through that state than others. The best state is the only one that gets better. 

To prevent this, we use regularization to cause the model to select each mode with uniform likelihood until the inference and emission network are well trained. Thus all discrete modes are able to learn the dynamics well initially. When the regularization decays, the transition dynamics of each mode can then specialize. One effect of this regularization strategy is that the weights for each dynamics module are correlated early during training and decorrelate when the regularization decays. The regularization helps the model to better utilize its capacity, and the model can achieve better likelihood, as demonstrated in~\Cref{sec:annealing} and~\Cref{fig:annealing}.

Multi-steps training has been used by previous models, and it serves the same purpose as our regularization. SVAE first trains a single transition model, then uses that one set of parameters to initialize all the transition dynamics for multiple states in next stage of training. rSLDS training begins by fitting a single AR-HMM for initialization, then fits a standard SLDS, before finally fitting the rSLDS model. We follow these implementations of both SVAE and rSLDS in our paper. Both multi-step training and our regularization ensure the hidden dynamics are well learned before learning the segmentation. What makes our regularization approach interesting is that it allows the model to be trained with a smooth transition between early and late training.

\end{document}